\documentclass[10pt,twocolumn,letterpaper]{article}

\usepackage{cvpr}              
\definecolor{cvprblue}{rgb}{0.21,0.49,0.74}
\usepackage[pagebackref,breaklinks,colorlinks,allcolors=cvprblue]{hyperref}

\def\paperID{} 
\def\confName{CVPR}
\def\confYear{2026}

\title{{\includegraphics[height=1.45em]{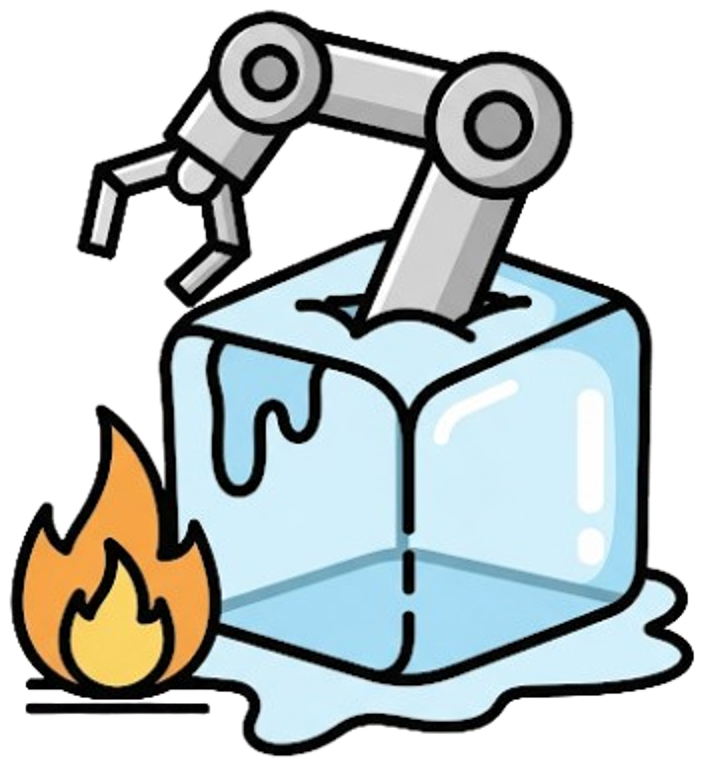}}
\textit{\textcolor{myskyblue}{Sta}\textcolor{myorange}{Mo}}: Unsupervised Learning of Generalizable Robot \textcolor{myorange}{Mo}tion from Compact \textcolor{myskyblue}{Sta}te Representation}

\vspace{-6mm}

\author{
Mingyu Liu$^{1,2,*}$ \quad
Jiuhe Shu$^{1,*}$ \quad
Hui Chen$^{1}$ \quad
Zeju Li$^{1}$ \quad
Canyu Zhao$^{1}$ \\
Jiange Yang$^{3}$ \quad
Shenyuan Gao$^{4}$ \quad
Hao Chen$^{1,\dagger}$ \quad
Chunhua Shen$^{1,5,\dagger}$ \\
$^1$State Key Laboratory of CAD \& CG, Zhejiang University \quad
$^2$Shanghai Innovation Institute \\
$^3$Nanjing University \quad
$^4$HKUST \quad
$^5$Ant Group \\
\tt\small mingyuliu@zju.edu.cn, chunhuashen@zju.edu.cn\\ 
}

\usepackage{float}
\usepackage{wrapfig}
\usepackage{stfloats}
\usepackage{newunicodechar}
\usepackage{multirow}
\newunicodechar{π}{\pi}
\definecolor{myskyblue}{RGB}{135, 206, 235}
\definecolor{myorange}{HTML}{FF8C00}
\def\NickName{\textit{\textcolor{myskyblue}{Sta}\textcolor{myorange}{Mo}}}

\begin{document}
\maketitle
\begingroup
\renewcommand\thefootnote{}
\footnotetext{$^*$Equal contribution.\quad $^\dagger$Corresponding authors.}
\endgroup
\begin{abstract}
A fundamental challenge in embodied intelligence is learning state representations that are both compact and expressive for world modeling and decision making, yet existing methods often remain either redundant or miss task-critical information. We propose an unsupervised approach that learns a highly compressed two-token state representation using a lightweight encoder and a pre-trained Diffusion Transformer (DiT) decoder, capitalizing on its strong generative prior. Our representation is efficient, interpretable, and integrates seamlessly into existing VLA-based models, improving performance by \textbf{11.6\%} on LIBERO and \textbf{31\%} in real-world task success rate with minimal inference overhead. More importantly, we find that the difference between these tokens, obtained via latent interpolation, naturally 
represents the motion, 
which can be further decoded into executable robot actions. This emergent capability reveals that our representation captures dynamics without explicit motion supervision. We name our method \textit{\NickName} for its ability to learn generalizable robotic \textit{\textcolor{myorange}{Mo}tion} from compact \textit{\textcolor{myskyblue}{Sta}te representation}, which is encoded from static images, \textit{challenging the prevalent dependence on learning 
robotic motions with complex temporal modeling and video data.} 
Our learned representations also enhance policy co-training, outperforming prior methods by \textbf{10.4\%} with improved interpretability. Moreover, our approach scales effectively across diverse data sources, including real-world robot data, simulation, and human egocentric video. Project page at \href{https://aim-uofa.github.io/StaMo/}{https://aim-uofa.github.io/StaMo/}.

\end{abstract}
  
\vspace{-2mm}
\section{Introduction}
\emph{``What we observe as static is merely dynamic equilibrium.''}\\
\hspace*{\fill} — Richard Feynman, \textit{The Feynman Lectures on Physics}
\vspace{0.5em}

\begin{figure*}[h]
\begin{center}
\includegraphics[width=.85\linewidth]{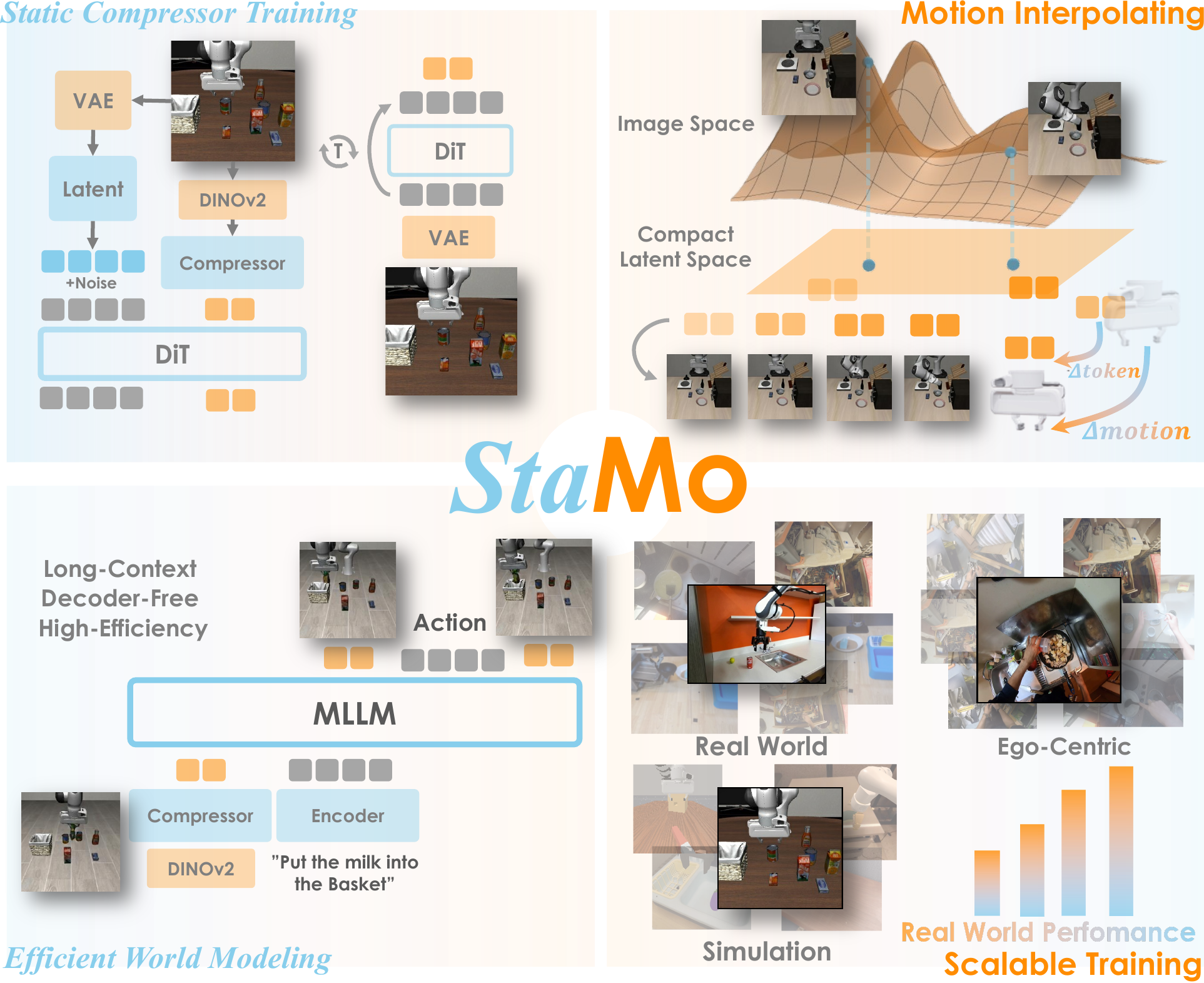} 
\end{center}
\caption{\textbf{An overview of our \textit{\NickName} framework.} Our method efficiently compresses and encodes robotic visual representations, enabling the learning of a compact state representation. Motion naturally emerges as the difference between these states in the highly compressed token space. This approach facilitates efficient world modeling and demonstrates strong generalization, with the potential to scale up with more data.}
\label{fig:teaser}
\vspace{-2mm}
\end{figure*}

Learning reusable and generalizable representations is a cornerstone of intelligent robotics systems. While visual representations in VLAs retain rich perceptual details for multimodal fusion, state representations for world modeling and intermediate reasoning serve a different purpose: they are designed to enable efficient future prediction and bridging visual planning with action execution. This role demands two key properties: \textit{\textbf{First}}, it must be highly compact. Unlike rich visual embeddings used for perception, a state representation's proximity to action generation demands a focus on actionable information to ensure efficient future prediction. \textit{\textbf{Second}}, it must be expressive despite its compactness. Common low-dimensional approaches, such as trajectories~\citep{atm} or flow fields~\citep{gao2024flip, flow}, often fail on this front; they capture basic motion but lack the semantic richness to encode goal states, interaction dynamics, or structured spatial relationships.

Building on these principles, we propose a compact state representation learned using a Diffusion Autoencoder fine-tuned on robotics data. Our method compresses an image into a highly compact latent state (as few as 2 tokens of 1024 dimensions). To ensure the representation is rich, we initialize its decoder from a powerful DiT model pretrained on internet data~\citep{esser2024scaling}, reasoning that the ability to reconstruct pixels accurately requires an implicit understanding of key state information like robot poses and object interactions. This representation integrates seamlessly into existing VLM architectures, extensive experiments show that our method significantly enhances model performance with negligible overhead to inference speed, achieving improvements of + 11.6\% in LIBERO and + 31\% in real-world deployments. 

Interestingly, from this compact state representation, we make an important discovery: robot motion (which is also known as latent action) naturally emerges from the state representation space. By simply performing a linear interpolation between the latent encodings of start and goal observations, we can generate smooth, plausible, and dynamically consistent motion trajectories. This observation provides an alternative to the dominant paradigm of learning latent actions from video data~\citep{lapa,bu2025univla,yang2025learning,moto}. While using video seems intuitive as action is inherently temporal, it presents significant challenges: it demands complex, computationally expensive temporal models and often produces ambiguous, coarse-grained actions. This is because high-variance motion within video clips causes models to learn averaged-out representations, making per-frame learning both inefficient and representationally flawed.

These challenges motivated us to revisit the fundamental motivation for using video data, \textbf{if the objective is to capture latent actions through frame-wise changes, why should we commit to learning complex motion extractors from suboptimal state representations?} We therefore raise a more fundamental question: \textbf{rather than explicitly modeling motion from sequences, can we learn a sufficiently expressive state representation from individual frames such that the simple difference between two states naturally encapsulates a meaningful latent action? }Our work demonstrates this is not only feasible but also highly advantageous. Compared to traditional video-based models, our state-centric approach is more training-efficient and avoids the representational ambiguity caused by motion variations in video. The emergent latent actions exhibit strong generalization and seamless sim-to-real transferability. We validate this effectiveness through extensive co-training experiments, where our latent actions achieve superior performance over traditional and more complex latent action learning approaches.

The captivating properties of state and motion ultimately form the foundation of our method, dubbed \textit{\NickName}, which learns generalizable \textit{\textcolor{myorange}{Mo}tion} from compact \textit{\textcolor{myskyblue}{Sta}te representation}. While the state is \textit{static}, the motion is \textit{dynamic}, a harmonious balance is elegantly achieved in our approach. We hope our method would shed new light on future research. Our contributions are summarized as follows:

\begin{itemize}[leftmargin=*]
\itemsep 0cm
\item We propose \textit{\NickName}, a novel framework that encodes a compact state representation from static images, from which motion naturally emerges.
\item Our representation can be efficiently utilized for world modeling and serves as an intermediate representation that bridges vision-language models (VLMs) and action expert modules. It can be seamlessly integrated into existing VLM-based frameworks, delivering improved performance with minimal inference overhead.
\item Our motion extraction approach offers enhanced flexibility, strong generalization capability, and excellent transferability. It can be effectively utilized for co-training downstream models and facilitating goal-image planning tasks.
\item Comprehensive simulation and real-world experiments, along with extensive visualizations, demonstrate the effectiveness of our approach, which can be readily scaled up with additional data.
\end{itemize}

\section{Related Works}

\subsection{Robot Representation Learning}


Previous robotics research on visual representations has often faced a fundamental trade-off: methods excelling at motion representation, like latent actions~\citep{bu2025univla,lapa,gr2,gr00t,gao2025adaworld}, flow~\citep{gao2024flip, xu2024flow}, or trajectories~\citep{atm}, typically lack a rich, compact state representation. Conversely, approaches that encode detailed state information from raw images~\citep{wang2025unified,gen2act} or dense features~\citep{zhang2025dreamvla,li2025bridgevla, nair2022r3m, xiao2022masked, majumdar2023we, radosavovic2023robot} are often too high-dimensional and redundant to effectively represent motion via simple differences. \textit{\NickName}, as shown in Figure~\ref{fig:where_is_stamo}, overcomes this dichotomy by achieving a unique balance. It learns to compress and encode a robot's visual state into a highly compact token space that is both expressive enough for complex tasks and minimal enough that motion can be elegantly and powerfully represented as the difference between two states. This unified approach not only enables efficient world modeling but also demonstrates superior generalization and scalability, providing a robust foundation for future robotic systems.


\subsection{World Modeling in VLA}

Benefiting from powerful vision-language models (VLMs)~\citep{bai2025qwen2,chen2024internvl,beyer2024paligemma} pre-trained on large-scale internet data, vision-language-action (VLA) models have demonstrated significant potential in generalizing across manipulation tasks. Typically, VLAs~\citep{brohan2022rt,zitkovich2023rt,cheang2024gr,ha2023scaling,kim2024openvla,black2410pi0,wen2025tinyvla,team2024octo,wang2025vq,liu2025bridge,wang2026odyssey} map visual information and language instructions into the robot’s action space through end-to-end training. A natural extension is to endow the model with world modeling capabilities. Inspired by recent unified generative and comprehension models, a promising direction is to enable large models to reason with images—predicting future visual states while inferring actions or language, thereby creating a mutually reinforcing process. Several works~\citep{wang2025unified,zhang2025dreamvla,cen2025worldvla,li2025unified} have explored this idea. However, these approaches either require decoding full images during inference or rely on overly redundant state representations, limiting their generalization to novel scenes. In contrast, our method employs a more compact representation that improves model performance with minimal impact on inference speed. Furthermore, our representation exhibits strong generalization ability, enabling adaptation to unseen scenes without requiring further fine-tuning of the encoder.

\subsection{Latent Action Learning}

Robot learning scalability is constrained by scarce and heterogeneous robotic data, prompting the use of large-scale, action-free internet videos to instill foundational physical and operational knowledge for improved generalization and data efficiency~\citep{survey}. Numerous studies have explored learning discrete~\citep{lapa,moto,genie,bu2025univla,lapo,agibot} or continuous~\citep{yang2025learning,gao2025adaworld} latent actions from unlabeled data and have demonstrated their effectiveness across a wide range of downstream tasks. However, most of these approaches rely on carefully designed model architectures and depend on extracting frames from continuous videos, which introduces limitations such as sensitivity to frame sampling intervals, potential temporal biases, and poorly interpretable, blurry action representations. In contrast, our approach reveals that actionable representations can inherently emerge from large-scale natural image collections—challenging the conventional belief that latent actions must be learned from video sequences. We further demonstrate that such image-emergent actions exhibit superior interpretability and stronger generalization capabilities.

\begin{figure}[h]
\begin{center}
\includegraphics[width=1\linewidth]{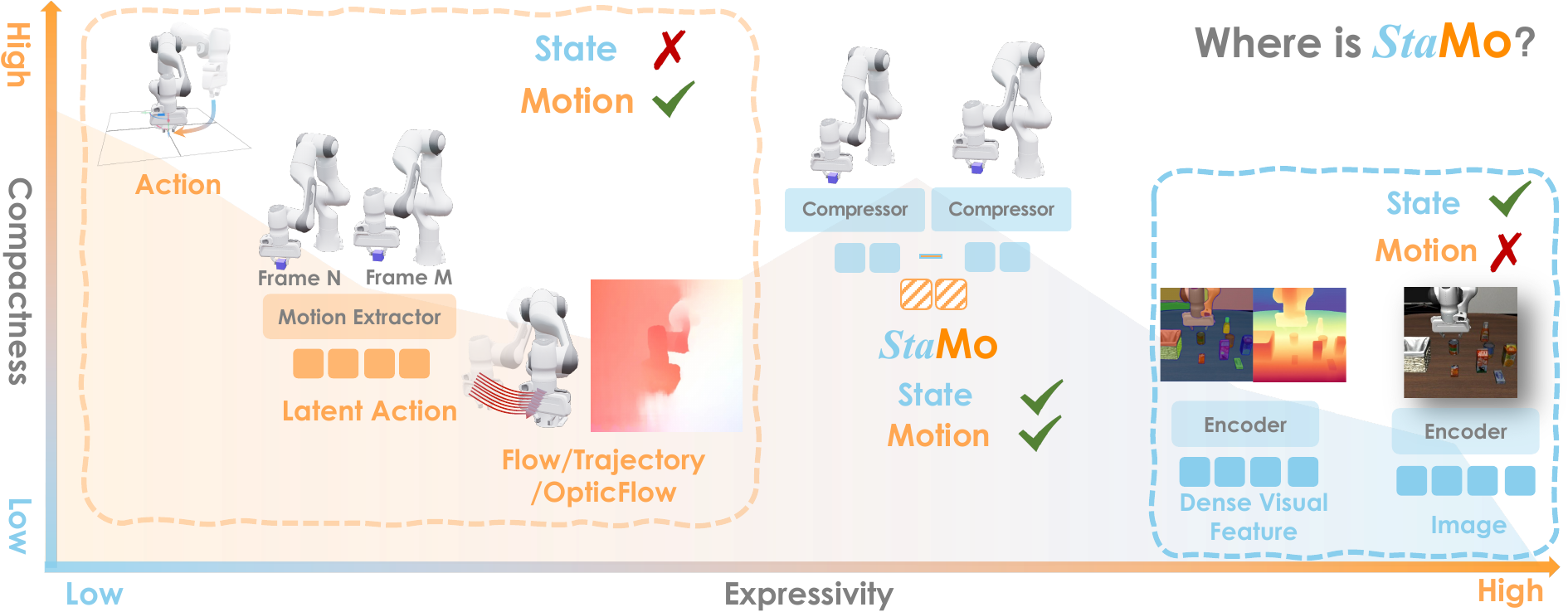} 
\end{center}
\caption{\textbf{Where is \textit{\NickName}?} This figure visualizes how different robotic representations fall on the spectrum of expressiveness versus compactness. StaMo uniquely occupies the ideal position, offering both a rich, expressive state representation and the ability to model motion from a highly compact space.}
\vspace{-5mm}
\label{fig:where_is_stamo}
\end{figure}

\section{Method}
\label{sec:method}

\subsection{Compress Image into Compact State}
A straightforward approach to state representation is to employ pretrained image encoders to process observations, using their output features as latent states. However, this strategy yields prohibitively large feature maps (e.g., 256$\times$1024), which contain significant redundant information and impose limitations on real-time execution and long-horizon planning in robotics. 
While using the single [CLS] token from the large feature maps substantially reduces dimensionality, this representation is often too coarse to facilitate effective and precise task execution. 
To resolve this trade-off between dimensionality and representational fidelity, we employ a Diffusion Autoencoder. We fine-tune this architecture specifically for robotic settings, enabling it to act as a powerful state compressor. This approach successfully compresses the high-dimensional observation into a highly compact latent state of as few as two 1024-dimensional tokens, capturing essential information while remaining computationally efficient.


To this end, as illustrated in Figure~\ref{fig:teaser}, the encoder of our Diffusion Autoencoder, denoted as $\mathcal E$, consists of a pretrained DINOv2 model~\citep{oquab2023dinov2} serving as a frozen feature extractor, which is followed by a transformer-based compressor, to map the input observations into a sequence of compact token representations. The decoder $\mathcal D$, is a Diffusion Transformer (DiT)~\citep{peebles2023scalable} conditioned on these tokens to reconstruct the original image observations. Our implementation is built upon the Stable Diffusion 3, where only the transformer-based compressor and the DiT decoder are trained. The DINOv2 encoder remains frozen throughout the training process.  We employ the same Flow Matching~\citep{albergo2022building,liu2022flow,lipman2022flow} objective used in base model training to optimize the model, which minimizes the loss between the velocity predicted by $\mathcal D$ and the target velocity $u$:
\begin{equation}
    \begin{aligned}
        \mathbf z_0 &= \tau(\mathbf x_0), \\
        { L_{DAE} } = 
        {\mathbb E}_{\mathbf z_0,t} \Vert \mathcal D&(\mathbf z_t,\mathcal{E}(\mathbf x_0),t) - u(\mathbf z_t) \Vert^2_2,
    \end{aligned}
\end{equation}
where $\mathbf x_0$ is the input image observation and $\tau$ is the VAE encoder of pretrained diffusion model to produce latent $\mathbf z_0$. $\mathbf z_t = (1-\sigma_t)\mathbf z_0 + \sigma_t \epsilon$ is the linear interpolation between pure noise $\epsilon$ and $\mathbf z_0$. After training, The trained encoder $\mathcal E$ is able to yield the state representation $s$. 


\subsection{Interpolating Motion from States}
A key advantage of our learned representation is its capacity to serve a dual role, representing both state and motion. This duality resolves a fundamental trade-off inherent in conventional robotics representations, which we analyze comparatively in Figure~\ref{fig:where_is_stamo}.
On one hand, traditional motion representations such as end-effector poses, optical flow, trajectories, or recent latent action models, are typically low-dimensional and compact. Their strength lies in capturing dynamics, often through simple differencing, making them effective for representing actions.
However, this compactness comes at the cost of expressivity: they lack the rich visual context necessary to reconstruct a plausible or detailed state.
On the other hand, prevailing state representations, derived from encoding raw image observations or supplementary inputs like depth and segmentation maps, capture rich semantic and geometric details. While highly expressive, these high-dimensional features are computationally burdensome and fail to intrinsically encode the dynamics or motion required to transition between states.
Our approach elegantly bridges this gap. By defining motion as the vector difference between consecutive compact state tokens $a_t = s_{t+1}-s_t$, we achieve a representation that is both compact and expressive, unifying the description of ``what the world looks like" and ``how the world changes".
The effectiveness of representing motions as the differences between states is demonstrated in our experiments in Section~\ref{sec:linear_probing}.

\subsection{StaMo for Efficient World Modeling}

Existing VLA models, such as our OpenVLA baseline, typically operate as reactive policies. They learn a direct mapping from the current visuolinguistic context to a sequence of low-level actions (e.g., 7-DoF end-effector controls). While effective, this paradigm does not explicitly compel the agent to reason about the physical consequences of its actions or how the world will change as a result. We hypothesize that endowing the agent with a predictive world model to anticipate future visual states will improve the future-planning capability of model. This auxiliary task of predicting ``what happens next" should, in turn, regularize the policy and improve the quality of the primary action prediction task.

The compact state representations produced by our StaMo encoder $\mathcal E$ are ideally suited for this purpose, which provide a concise yet rich summary of the environment for the model to reason about the future. To this end, we integrate the compact representations of StaMo into the OpenVLA architecture and train the model to jointly predict the next state and the corresponding action. Technically, we achieve this by attaching a lightweight MLP head to OpenVLA's autoregressive backbone, which is tasked specifically with predicting the subsequent state representation. The model is optimized with a composite loss function that balances action generation and future-state prediction:
\vspace{-1mm}
{
\small
\begin{multline}
    L_{total} = \lambda_{action}L_{action}  
              + \lambda_{future}(L_{mse}(s_{pred},s_{gt}) \\+ L_{1}(s_{pred},s_{gt})),
\end{multline}
}

where $L_{action}$ is the standard cross-entropy loss for next-token prediction, anchoring the model to its primary control task. The world model objective combines Mean Squared Error (MSE) and L1 losses to enforce accurate regression of the ground-truth future state. In our experiments, we set $\lambda_{action}=\lambda_{future}$, reflecting our hypothesis that learning to predict is as crucial as learning to act. Our experimental results validate this hypothesis, demonstrating that compelling the VLA to predict future states significantly improves task success rates, as detailed in Section~\ref{sec:world_modeling_results}.

\subsection{StaMo for Latent Motions Co-Training}



While Section~\ref{quality} provides a qualitative visualization of the emergent motion, the abstract nature of our latent motion representation makes direct quantitative comparison challenging. To rigorously assess its effectiveness, we employ a policy co-training strategy. This approach evaluates the utility of our motion representation on downstream tasks by jointly training a policy model on a small set of action-labeled robot data alongside a larger set of action-less video data. We infer a latent motion, $m_t$, for each pair of consecutive video frames by calculating the difference between their compact state representations encoded by our model $E(\cdot):
m_t = E(o_{t+1}) - E(o_t)$,
This process generates pseudo-action labels for the video data, creating a unified dataset where our emergent motions and ground-truth robot actions are learned jointly within a single policy model. This approach unlocks the potential to learn from vast and diverse video sources without requiring explicit action labels.

We tested this co-training framework in Section~\ref{sec:cotraining}, our method leads to substantial performance gains compared with previous method. The results confirm that the motions emerging from \textit{\NickName} serve as an effective and robust proxy for true actions, providing a scalable pathway to enhance policy learning by leveraging the wealth of unlabeled visual data.
\section{Experiments}

\subsection{Qualitative Analysis of StaMo}
\label{quality}

\vspace{-1mm}
 \textit{\NickName} demonstrates the ability to perform high-quality reconstructions of robotic manipulation images with strong generalization capabilities. It achieves excellent results on both in-domain and out-of-domain data. Quantitative reconstruction results are shown in Table~\ref{tab:reconstruc_metrics}, and some qualitative results are presented in Figure~\ref{fig:reconstruct}. More results can be found in Appendix. For main experiments, we use DROID~\citep{droid} and LIBERO~\citep{libero} as our training data.

\begin{table}[htbp]
\renewcommand{\arraystretch}{1.2}
\caption{Reconstruct Performance comparison of different datasets using our \textit{\NickName} encoder.}
\vspace{-2mm}
\centering
\footnotesize
\scalebox{0.55}{
\begin{tabular}{lccccccc}
\toprule
\textbf{Dataset} & \textbf{libero\_10} & \textbf{libero\_90} & \textbf{libero\_goal} & \textbf{libero\_object} & \textbf{libero\_spatial} & \textbf{Droid} & \textbf{Maniskill(OOD)}\\
\midrule
PSNR (dB) & 25.5194 & 27.2470 & 24.6467 & 27.0011 & 25.9683 & 20.2492 & 22.1673\\
SSIM     & 0.8909 & 0.8962 & 0.8926 & 0.9105 & 0.8984 & 0.7346 & 0.8824\\
\bottomrule
\end{tabular}
}
\label{tab:reconstruc_metrics}
\end{table}

Figure~\ref{fig:reconstruct} also showcases \textit{\NickName}'s powerful motion interpolation ability. By linearly interpolating the tokens in the latent space, the decoded images exhibit excellent continuity and plausible motion. Furthermore, we demonstrate the results of motion transfer. By taking the difference between tokens in the latent space, the resulting latent motion proves to be highly transferable—both in sim-to-sim, sim-to-real, and real-to-sim scenarios. This indicates that the learned motion is not scene-specific but possesses strong generalization capabilities.

\begin{figure}[!htp]
    \begin{center}
    \includegraphics[width=1\linewidth]{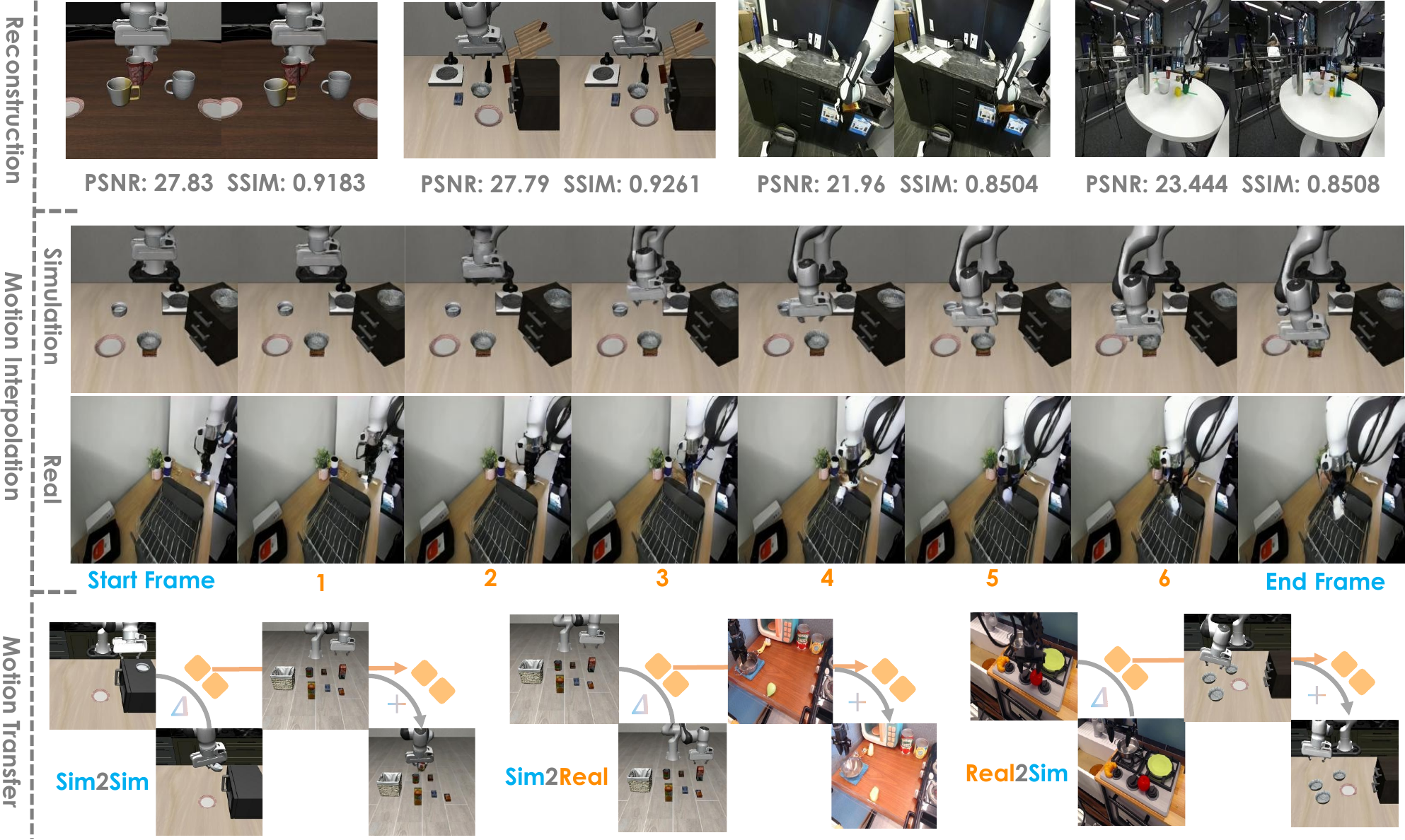} 
    \end{center}
    \vspace{-2mm}
    \caption{\textbf{Reconstruct images using our \textit{\NickName} encoder with as few as two 1024-dimensional tokens.} The first row shows the ground truth, and the second row shows the predicted results, with corresponding PSNR and SSIM metrics listed below. The results demonstrate that \textit{\NickName} can preserve high image fidelity and structural similarity even under extremely compressed state representations.}
    \label{fig:reconstruct}
    \vspace{-4mm}
\end{figure}

\subsection{World Modeling Results}
\label{sec:world_modeling_results}

In this section, we evaluate the application of our \textit{\NickName} representation for efficient world modeling. A key advantage of our approach is its ability to integrate seamlessly into existing VLM-based frameworks. We demonstrate this by conducting experiments on two strong baselines: OpenVLA and its successor, OpenVLA-OFT. We specifically chose OpenVLA-OFT because it introduces architectural improvements like parallel decoding and action chunking, which result in a significantly longer action horizon. This requires the model to predict a state representation at a more distant future step, making it a more challenging and relevant benchmark for world modeling. 

To validate our approach, we retrained both OpenVLA and OpenVLA-OFT, replacing their original vision encoders with our \textit{\NickName} encoder and training them to predict \textit{\NickName} states. We denote these modified architectures as OpenVLA* and OpenVLA-OFT*; the original, unmodified models are presented without the asterisk. For our OpenVLA-OFT* variant, we introduced a minor architectural modification, using only the third-person camera view and removing the wrist camera input. We evaluate the performance of our \textit{\NickName} representation on the LIBERO benchmark to demonstrate its superior capability in assisting VLA manipulation through enhanced world modeling. For each task, we conduct 1,000 evaluation rollouts, and all results are presented as percentage success rates.

\begin{table}[htbp]
    \renewcommand{\arraystretch}{1.2}
    \caption{Performance comparison of different methods (\%).}
    \vspace{-2mm}
    \centering
    \footnotesize
    \scalebox{.76}
    {
        \begin{tabular}{lccccc}
        \toprule
        \textbf{Method} & \textbf{Spatial} & \textbf{Object} & \textbf{Goal} & \textbf{Long} & \textbf{Average} \\
        \midrule
        DP~\citep{chi2023diffusion} & 78.3\% & 92.5\% & 68.3\% & 50.5\% & 72.4\% \\
        Octo-Base~\citep{team2024octo} & 78.9\% & 85.7\% & 84.6\% & 51.1\% & 75.1\% \\
        Spatial-VLA~\citep{qu2025spatialvla0} & 88.2\% & 89.9\% & 78.6\% & 55.5\% & 78.1\% \\
        CoT-VLA~\citep{zhao2025cot} & 87.5\% & 91.6\% & 87.6\% & 69.0\% & 83.9\% \\
        $\pi$0-FAST~\citep{pertsch2025fast} & 96.4\% & 96.8\% & 88.6\% & 60.2\% & 85.5\% \\
        WorldVLA~\citep{cen2025worldvla} & 87.6\% & 96.2\% & 83.4\% & 60.0\% & 81.8\% \\
        UniVLA~\citep{wang2025unified} & 95.4\% & 98.8\% & 93.6\% & 94.0\% & 96.5\%\\
        \midrule
        OpenVLA~\citep{kim2024openvla} & 84.7\% & 88.4\% & 79.2\% & 53.7\% & 76.5\% \\
        OpenVLA* + DINOv2 Feature & 88.6\% & 90.4\% & 83.5\% & 61.4\% & 80.9\% \\
        OpenVLA* + \textit{\NickName} state & 92.3\% & 94.8\% & 88.1\% & 75.2\% & 87.6\% \\
        OpenVLA* + \textit{\NickName} motion & 93.1\% & 95.1\% & 87.4\% & 76.9\% & 88.1\% \\
        \midrule
        OpenVLA-OFT~\citep{vla_improve} & 93.7\% & 94.2\% & 89.7\% & 91.3\% & 92.2\% \\
        OpenVLA-OFT* + DINOv2 Feature & 94.1\% & 95.0\% & 91.2\% & 93.0\% & 93.3\% \\
        OpenVLA-OFT* + \textit{\NickName} state & \textbf{96.8\%} & \textbf{98.9\%} & \textbf{95.0\%} & \textbf{96.3\%} & \textbf{96.8\%}\\
        OpenVLA-OFT* + \textit{\NickName} motion & 95.3\% & 97.1\% & 94.1\% & 92.9\% & 94.9\%\\
        \bottomrule
        \end{tabular}
    }
    \label{tab:performance_comparison}
\end{table}

Both the \textit{\NickName} state and motion representations can serve as predictive targets for a world model, and their performance is detailed in Table~\ref{tab:performance_comparison}. Our findings indicate that the optimal choice of representation depends on the policy's action horizon. For the standard OpenVLA, which predicts actions over a short, single-step interval, the motion representation yields better performance. This is because the latent motion is conceptually analogous to the delta end-effector (EE) pose used in frameworks like LIBERO, providing a direct and incremental target. Conversely, for OpenVLA-OFT, which plans over a longer execution step, the state representation is recommended, as it acts as a stable goal-conditioning signal.

\subsection{Inference Speed Comparison}

A key advantage of \textit{\NickName} lies in its utilization of two highly compressed tokens to assist VLM-based models with world modeling. During the inference stage, it obviates the need to decode full images; the model only needs to predict the tokens themselves. This approach significantly enhances the efficiency of the world modeling process. We conducted a comparative analysis of the model's inference frequency before and after integrating the \textit{\NickName} representation, and also benchmarked its speed against other world modeling methods. Our findings indicate that \textit{\NickName} introduces significantly less inference overhead. The comparison of different model's inference frequency is presented in Table~\ref{tab:frequency_comparison}.

\begin{table}[htbp]
    \centering
    \small
    \setlength{\tabcolsep}{6pt}
    \caption{Inference Speed Comparison.The table presents a comparison of inference speeds for different methods, with all models deployed on a single RTX 4090 GPU. We benchmark against other world model-based VLA approaches, such as UniVLA~\citep{wang2025unified} and WorldVLA~\cite{cen2025worldvla}. As the results show, our method achieves a higher inference frequency compared to previous methods, including directly using DINOv2 features.Our approach introduces minimal computational overhead compared to the original baseline model.}
    \vspace{-2mm}
    \label{tab:frequency_comparison}
    \scalebox{0.8}{
    \begin{tabular}{lc}
        \toprule
        \textbf{Method} & \textbf{Frequency (Hz)} $\uparrow$\\
        \midrule
        WorldVLA~\citep{cen2025worldvla} & 2.27 \\
        UniVLA~\citep{wang2025unified} & 2.65 \\
        OpenVLA~\citep{kim2024openvla} & 4.16 \\
        OpenVLA + DINOv2 Feature & 2.86 \\
        OpenVLA + \textit{\NickName} & 4.02 \\
        OpenVLA-OFT~\citep{vla_improve} & 18.24 \\
        OpenVLA-OFT + DINOv2 Feature & 11.42 \\
        OpenVLA-OFT + \textit{\NickName} & 17.82 \\
        \bottomrule
    \end{tabular}}
\end{table}

\begin{table}[htbp]
    \renewcommand{\arraystretch}{1.2}
    \caption{Zero-Shot Performance on Maniskill~\citep{gu2023maniskill2}.}
    \vspace{-2mm}
    \centering
    \footnotesize
    \scalebox{.75}
    {
        \begin{tabular}{lccccc}
        \toprule
        \textbf{Method} & \textbf{Pick Cube} & \textbf{Push Cube} & \textbf{Stack Cube} & \textbf{Pull Cube Tool} \\
        \midrule
OpenVLA~\citep{kim2024openvla} & 28\% & 4\% & 18\% & 16\%  \\
UniVLA~\citep{wang2025unified} & 32\% & 7\% & 22\% & 20\%  \\
WorldVLA~\citep{cen2025worldvla} & 30\% & 10\% & 17\% & 22\% \\
OpenVLA + \textit{\NickName} state & \textbf{41\%} & \textbf{32\%} & \textbf{48\%} & \textbf{36\%} \\
        \bottomrule
        \end{tabular}
    }
    \label{tab:maniskill}
\end{table}

\subsection{Zero-Shot Transfer to New Environment}
Due to the compact nature of the \textit{\NickName} representation, which focuses on manipulation-relevant information rather than extraneous visual details, our model exhibits superior generalization capabilities. We validated this in Table~\ref{tab:maniskill} by taking the model trained on the LIBERO benchmark and applying it directly to the ManiSkill~\cite{gu2023maniskill2} environment. Crucially, all methods were deployed in this zero-shot transfer setting. We evaluated each task over 100 trials, with results presented as percentage success rates. The outcomes demonstrate that our approach achieves stronger performance than competing models under these conditions.

\subsection{Policy Co-training Results}
\label{sec:cotraining}

In Table~\ref{tab:rdt_performance}, we detail the results of our policy co-training. Our approach utilizes a DIT-based model on the RDT architecture, trained on a per-task dataset of 10 robot trajectories and 40 video trajectories with StaMo-generated pseudo-actions. We evaluated the final model checkpoint for each task across 20 trials, averaging the success rate over three full runs. StaMo significantly outperforms both LAPA~\citep{lapa} and ATM~\citep{atm}, suggesting that its pseudo-action labels are more faithful to the original actions.

\begin{table}[htbp]
    \renewcommand{\arraystretch}{1.2}
    \caption{Performance comparison of RDT with different data configurations (\%).}
    \vspace{-2mm}
    \centering
    \footnotesize
    \scalebox{.85}
    {
        \begin{tabular}{lccccc}
        \toprule
        \textbf{Method} & \textbf{Spatial} & \textbf{Object} & \textbf{Goal} & \textbf{Long} & \textbf{Average} \\
        \midrule
        RDT (all Real) & 91.6\% & 93.3\% & 86.7\% & 73.3\% & 86.2\% \\
        RDT (1 Real) & 71.7\% & 70.0\% & 66.7\% & 43.3\% & 62.9\% \\
        RDT (1Real+ 4ATM~\citep{atm}) & 83.3\% & 81.7\% & 71.7\% & 56.7\% & 73.4\% \\
        RDT (1Real+ 4LAPA~\citep{lapa}) & 80.0\% & 76.7\% & 75.0\% & 65.0\% & 74.2\% \\
        RDT (1Real + 4\textit{\NickName}) & \textbf{90.0\%} & \textbf{91.6\%} & \textbf{86.7\%} & \textbf{70.0\%} & \textbf{84.6\%}\\
        \bottomrule
        \end{tabular}
    }
    \label{tab:rdt_performance}
\end{table}

\subsection{Action Linear Probing Results}
\label{sec:linear_probing}
A central claim of our work is that an effective latent action can be formulated as the simple vector difference between two latent states encoded by \textit{\NickName}. To quantitatively validate the quality and utility of these emergent latent actions, we employ a linear probing  protocol~\citep{alain2016understanding}, a standard methodology for evaluating the efficacy of learned representations, to measure how much explicit, predictable information about the motion is contained within our latent action formulation.

We first construct a dataset by randomly sampling tuples from a large collection of robot trajectories. Each sampled tuple is of the form $\left( \mathbf{I}_n, \mathbf{I}_{n+k}, \mathbf{A}_n \right)$, where $\mathbf{I}_n$ is a starting image, $\mathbf{I}_{n+k}$ is the goal image after a horizon of $k$ steps, and $\mathbf{A}_n = \left( \mathbf{a}_n, …,\mathbf{a}_{n+k-1}  \right)$ is the sequence of ground-truth robot actions executed between them. Each image pair is then encoder into latent action, $\Delta z$, using our frozen \textit{\NickName} encoder, $\mathcal{E}$:

\vspace{-5mm}

\begin{equation}
    \Delta z = \mathcal{E}\left(\mathbf{I}_{n+k}\right) - \mathcal{E}\left(\mathbf{I}_n\right).
\end{equation}

\vspace{-2mm}

We then train a lightweight Multi-Layer Perceptron (MLP) to predict the action sequence $\mathbf{A}_n$ from the latent action representation $\Delta z$. The performance is measured by the Mean Squared Error (MSE) between the predicted action sequence $\hat{\mathbf{A}}_n$ and the ground-truth sequence $\mathbf{A}_n$. This process can be formally described as follows.

\vspace{-5mm}

\begin{equation}
\operatorname{MSE}\left( \mathbf{A}_n, \mathbf{\hat{A}}_n\right) = \operatorname{MSE} \left(\mathbf{A}_n, \operatorname{MLP}(\Delta z) \right).
\end{equation}

\vspace{-2mm}

A low LP-MSE score provides strong quantitative evidence that the vector difference between our latent states serves as a highly informative and linearly separable representation for robotic motions.



\vspace{-2mm}

\begin{figure}[h!]
    \centering
    \includegraphics[width=0.48\textwidth]{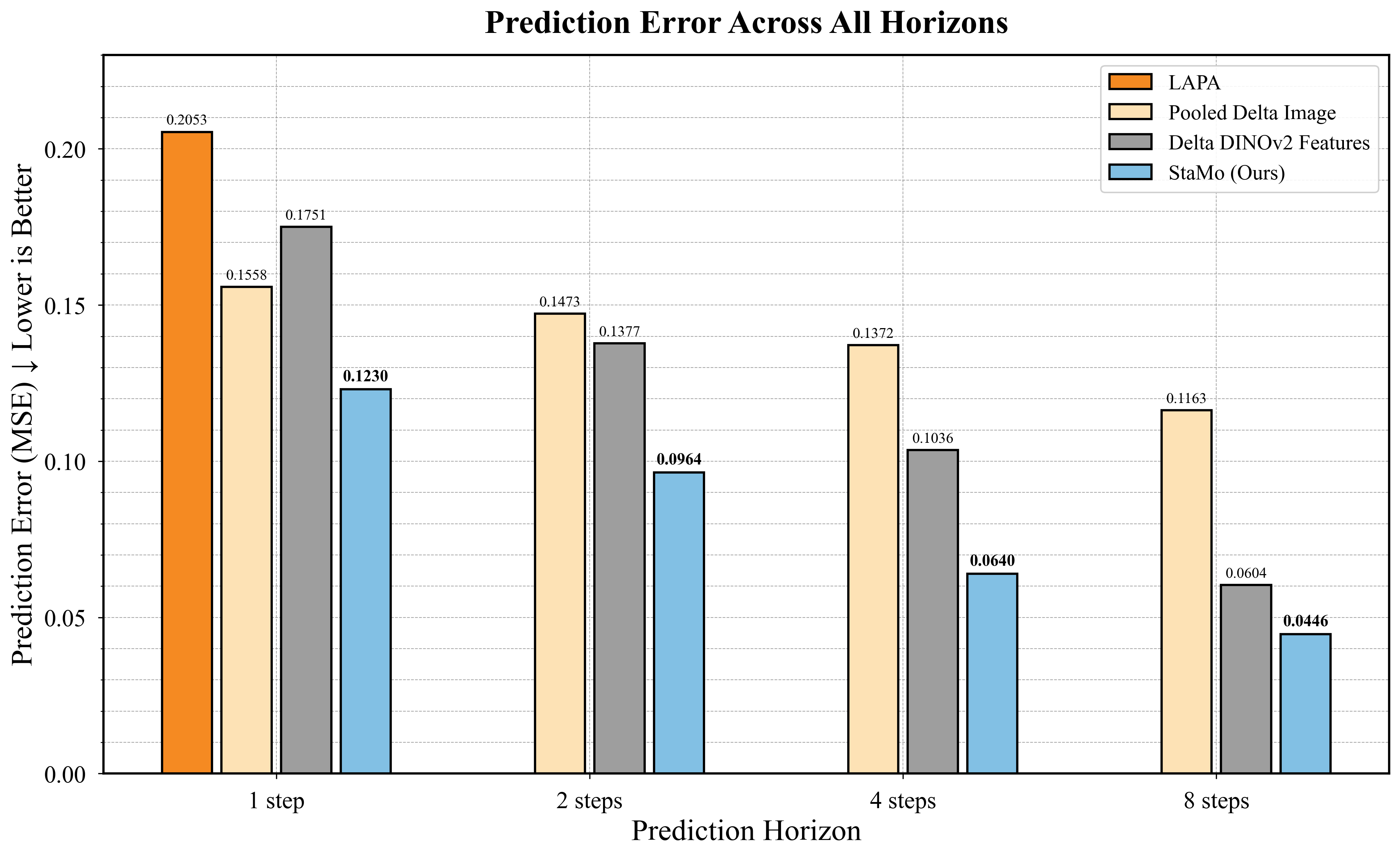}
    \vspace{-5mm}
    \caption{\textbf{Linear Probing MSE results.} We compare our method against three baselines. Our method consistently achieves the \textbf{lowest} MSE across all horizons.}
    \label{fig:LP-MSE}
\end{figure}

\vspace{-3mm}

To validate our approach, we benchmark our latent action representation against several baselines. The first set of baselines tests the importance of our structured latent space against alternatives operating on raw data: Pooled Delta Image (using pixel-wise differences) and Delta DINOv2 Features (using feature-space differences). The second baseline, LAPA~\citep{ye2024latent}, contrasts our deterministic state-difference formulation with a state-of-the-art autoregressive model that generates latent action tokens.

For a fair comparison, all representations are used to train an identical lightweight MLP. We evaluate non-generative methods across 1, 2, 4, and 8-step prediction horizons, while LAPA is assessed at its native 1-step horizon. Once trained, the MLP can directly map a current observation and a goal image to an executable action trajectory on the test set. The results in Figure~\ref{fig:LP-MSE} confirm the superiority of our method.


\subsection{Goal-Conditioned Task Planning}

Once the linear head is trained, it becomes capable of translating abstract representations into concrete, executable actions. We evaluated the goal-conditioned task planning capabilities of \textit{\NickName} and DINO-WM~\citep{zhou2024dino} using the ManiSkill benchmark. ManiSkill was specifically selected because it was not included in the pre-training datasets for either \textit{\NickName} or DINO-WM. For this comparison, we trained only the linear head while keeping the encoders of both \textit{\NickName} and DINO-WM frozen. Table~\ref{tab:dino_wm} presents the performance comparison, and further implementation details are available in the appendix. We evaluated each task over 100 trials, with results presented as percentage success rates.
\begin{table}[htbp]
    \renewcommand{\arraystretch}{1.2}
    \caption{Goal-Conditioned Task Planning.}
    \vspace{-2mm}
    \centering
    \footnotesize
    \scalebox{.75}
    {
        \begin{tabular}{lccccc}
        \toprule
        \textbf{Method} & \textbf{Pick Cube} & \textbf{Push Cube} & \textbf{Stack Cube} & \textbf{Pull Cube Tool} \\
        \midrule
DINO-WM~\citep{zhou2024dino} & 12\% & 8\% & 4\% & 18\% \\
\textit{\NickName} & \textbf{22\%} & \textbf{32\%} & \textbf{16\%} & \textbf{38\%} \\
        \bottomrule
        \end{tabular}
    }
    \label{tab:dino_wm}
\end{table}

\vspace{-2mm}

\subsection{Real World Experiments}

\begin{table*}[htbp]
\renewcommand{\arraystretch}{1.2}
\caption{Performance comparison on short and long horizon tasks.Best results are in \textbf{bold} and second-best results are underlined.}
\vspace{-2mm}
\centering
\footnotesize
\scalebox{0.85}
{
\begin{tabular}{lccccccccc}
\toprule
& \multicolumn{3}{c}{\textbf{Short horizon tasks}} & & \multicolumn{3}{c}{\textbf{Long horizon tasks}} & & \\
\cmidrule(lr){2-4} \cmidrule(lr){6-8}
\textbf{Method} & \begin{tabular}{@{}c@{}}Pick up \\the [toyname]\end{tabular} & \begin{tabular}{@{}c@{}}put the toy \\into the basket\end{tabular} & \begin{tabular}{@{}c@{}}Open the \\drawer\end{tabular}  & \textbf{Average} & \begin{tabular}{@{}c@{}}put all cups \\into the basket\end{tabular} & \begin{tabular}{@{}c@{}}put the toy \\into the drawer\end{tabular} &  \begin{tabular}{@{}c@{}}Stack all \\cups\end{tabular} & \textbf{Average} & \begin{tabular}{@{}c@{}}All \\Average\end{tabular} \\
\midrule
Spatial-VLA~\citep{qu2025spatialvla0} & 0.45 & 0.50 & 0.20 & 0.38 & 0.35 & 0.20 & 0.20 & 0.25 & 0.32 \\
WorldVLA~\citep{cen2025worldvla} & 0.55 & 0.20 & 0.15 & 0.30 & 0.25 & 0.40 & 0.15 & 0.26 & 0.28 \\
UniVLA~\citep{wang2025unified} & 0.55 & 0.30 & 0.35 & 0.40 & 0.35 & 0.30 & 0.10 & 0.25 & 0.33 \\
OpenVLA~\citep{kim2024openvla} & 0.35 & 0.30 & 0.25 & 0.30 & 0.20 & 0.30 & 0.15 & 0.20 & 0.25 \\
OpenVLA + \textit{\NickName} state & \underline{0.60}& \underline{0.55} & \underline{0.50} & \underline{0.60} & \underline{0.55} & \underline{0.55} & \underline{0.45} & \underline{0.52} & \underline{0.56} \\
OpenVLA-OFT + \textit{\NickName} state & \textbf{0.70} & \textbf{0.65} & \textbf{0.60} & \textbf{0.65} & \textbf{0.70} & \textbf{0.65} & \textbf{0.55} & \textbf{0.63} & \textbf{0.64} \\
\bottomrule
\end{tabular}
}
\label{tab:horizon_tasks}
\end{table*}

\begin{figure}[h]
\vspace{-3mm}
\begin{center}
\includegraphics[width=1\linewidth]{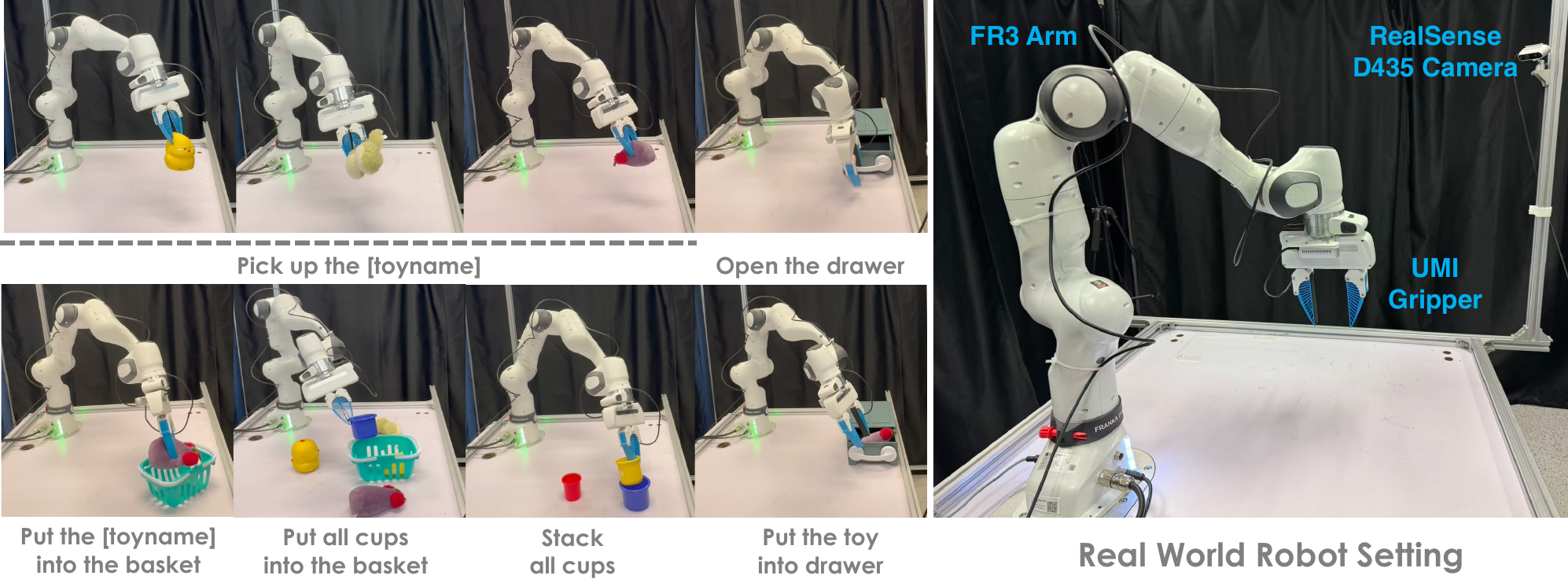} 
\end{center}
\vspace{-5mm}
\caption{\textbf{Real World Setting and Tasks.} We designed a benchmark of six real-world robotics tasks, spanning both short- and long-horizon challenges, to evaluate the effectiveness of our representation for learning world models.}
\label{fig:real}
\end{figure}

\textbf{Experiment Setup} Our real-world experimental benchmark comprises six manipulation tasks (three short-horizon and three long-horizon) to comprehensively evaluate the effectiveness of our \textit{\NickName} representation in facilitating world modeling and decision-making across a spectrum of task complexities, as shown in Figure~\ref{fig:real}. For each task, we collected 50 human demonstrations and evaluated performance over 20 trials. The results, presented as percentage success rates, are shown in Table~\ref{tab:horizon_tasks}.

\vspace{-2mm}
\subsection{Ablation Study}
\subsubsection{Ablation of Token Dimension} 
We ablate the hidden dimension of \NickName’s 2-token latent state (trained on LIBERO datasets, evaluated via PSNR/SSIM). As shown in Table~\ref{tab:hidden_dims}, across 256/512/1024 dimensions. This indicates hidden dimension has limited impact on performance, as the pre-trained DiT decoder’s generative prior offsets potential information loss in lower dimensions. 



\begin{table}[h]
\renewcommand{\arraystretch}{1.2}
\caption{Reconstruction performance with \textbf{2 tokens} under different hidden dimensions using our \textit{\NickName} encoder.}
\label{tab:hidden_dims}
\vspace{-2mm}
\centering
\footnotesize
\scalebox{0.65}{
\begin{tabular}{ccccccc}
\toprule
\textbf{hidden\_dim} & \textbf{Datasets} & \textbf{libero\_10} & \textbf{libero\_90} & \textbf{libero\_goal} & \textbf{libero\_object} & \textbf{libero\_spatial} \\
\midrule
\multirow{2}{*}{256}
 & PSNR (dB) & 26.5969 & 28.5858 & \textbf{27.5070} & \textbf{31.2779} & 25.7993 \\
 & SSIM      & 0.9082  & 0.8948  & \textbf{0.9327}  & 0.9478           & 0.9118 \\
\midrule
\multirow{2}{*}{512} 
 & PSNR (dB) & \textbf{27.3322} & \textbf{29.5773} & 26.2000 & 31.0129 & 24.6279 \\
 & SSIM      & 0.9180           & \textbf{0.9127}  & 0.9194  & 0.9444  & 0.9009 \\
\midrule
\multirow{2}{*}{1024} 
 & PSNR (dB) & 27.0418 & 29.4810 & 26.4807 & 31.2653 & \textbf{25.9964} \\
 & SSIM      & \textbf{0.9186} & 0.9124 & 0.9194 & \textbf{0.9480} & \textbf{0.9155} \\
\bottomrule
\end{tabular}
}
\vspace{-5mm}
\end{table}

\subsubsection{Ablation of Additional Training Data}

\begin{figure}[htbp]
\vspace{-5mm}
    \centering
    \includegraphics[width=0.4\textwidth]{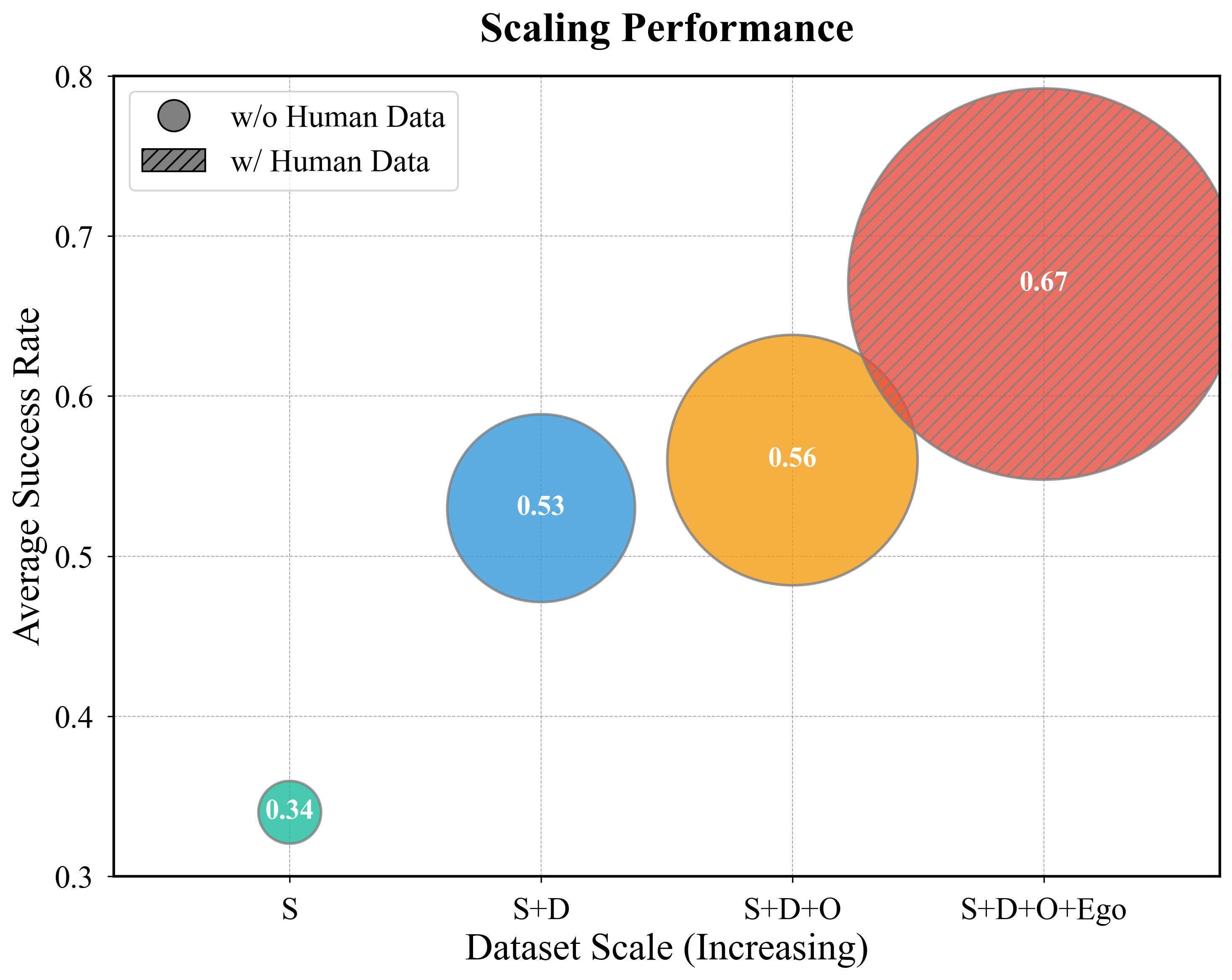}
    
    \caption{\textbf{Scaling Performance.} Performance of our model can be scale with more data, including human ego-centric data.``S" means Simulation data only,``D" represents DROID~\citep{droid} data, ``O" represents OXE~\citep{openx} data, ``Ego" means Ego-centric data.}
    \label{fig:scale}
\end{figure}

A natural question following our initial validation is the scalability of our approach. To explore this, we investigated the effect of the \textit{\NickName} training data volume and diversity on real-world task performance. We compared several data configurations: (1) simulation data only, (2) simulation + DROID~\citep{droid} data (our main setting), (3) simulation + DROID + OXE~\cite{openx} data, and (4) simulation + DROID + OXE + egocentric human (Ego) data~\citep{egovid}. The results, clearly depicted in Figure~\ref{fig:scale}, confirm that our model is highly scalable. This verifies that incorporating larger volumes of data and greater task diversity, such as varied robot embodiments from OXE and human-centric data, steadily improves \textit{\NickName}'s performance capabilities.

\subsubsection{Ablation of Pretrained Diffusion Decoder}

We compared the performance of using a pre-trained diffusion decoder against training one entirely from scratch. The decoder trained from scratch required a significantly longer time to converge and ultimately was unable to match the same level of performance as its pre-trained counterpart. As shown in Table~\ref{tab:pretrain_ablation}. This finding demonstrates the critical importance of pre-training on large-scale natural image datasets. Such pre-training provides a crucial prior, which is highly beneficial for learning a more effective representation of the robotics data distribution.

\begin{table}[htbp]
    \renewcommand{\arraystretch}{1.2}
    \caption{Performance comparison between using pretrained diffusion decoder and training from scratch.}
    \vspace{-2mm}
    \centering
    \footnotesize
    \scalebox{.75}
    {
        \begin{tabular}{lccccc}
        \toprule
        \textbf{Method} & \textbf{Spatial} & \textbf{Object} & \textbf{Goal} & \textbf{Long} & \textbf{Average} \\
        \midrule
        Train from Scratch & 90.3\% & 92.7\% & 86.5\% & 73.4\% & 85.7\% \\
        w. Pretrain Diffusion Decoder & \textbf{92.3\%} & \textbf{94.8\%} & \textbf{88.1\%} & \textbf{75.2\%} & \textbf{87.6\%} \\
        \bottomrule
        \end{tabular}
    }
    \label{tab:pretrain_ablation}
\end{table}
\vspace{-5mm}

\section{Conclusion}
In this work, we introduced \textit{\NickName}, a framework that extracts reusable latent action vectors  from static images via a transformer-based tokenizer and a diffusion decoder. We demonstrate that linear interpolation in this latent space produces smooth, physically plausible motions, and that these vectors generalize zero-shot to unseen senarios, suggesting that large-scale visual models implicitly learn a linearized dynamics manifold. Our approach offers a pathway toward scalable unsupervised skill discovery from diverse visual data, bridging the gap between static perception and dynamic action.
\clearpage
\section*{Acknowledgments}
{\it  This work was supported in part by 
the Zhejiang Provincial Pioneering Science and Technology Program (2025), and by the National Natural Science Foundation of China (No. 62576315).
}

{
    \small
    \bibliographystyle{ieeenat_fullname}
    \bibliography{main}

@String(AAAI = {AAAI})

@article{brohan2022rt,
  title={Rt-1: Robotics transformer for real-world control at scale},
  author={Brohan, Anthony and Brown, Noah and Carbajal, Justice and Chebotar, Yevgen and Dabis, Joseph and Finn, Chelsea and Gopalakrishnan, Keerthana and Hausman, Karol and Herzog, Alex and Hsu, Jasmine and others},
  journal={arXiv preprint arXiv:2212.06817},
  year={2022}
}

@inproceedings{zitkovich2023rt,
  title={Rt-2: Vision-language-action models transfer web knowledge to robotic control},
  author={Zitkovich, Brianna and Yu, Tianhe and Xu, Sichun and Xu, Peng and Xiao, Ted and Xia, Fei and Wu, Jialin and Wohlhart, Paul and Welker, Stefan and Wahid, Ayzaan and others},
  booktitle={Conference on Robot Learning},
  pages={2165--2183},
  year={2023},
  organization={PMLR}
}

@article{lipman2022flow,
  title={Flow matching for generative modeling},
  author={Lipman, Yaron and Chen, Ricky TQ and Ben-Hamu, Heli and Nickel, Maximilian and Le, Matt},
  journal={arXiv preprint arXiv:2210.02747},
  year={2022}
}

@article{liu2022flow,
  title={Flow straight and fast: Learning to generate and transfer data with rectified flow},
  author={Liu, Xingchao and Gong, Chengyue and Liu, Qiang},
  journal={arXiv preprint arXiv:2209.03003},
  year={2022}
}

@article{oquab2023dinov2,
  title={Dinov2: Learning robust visual features without supervision},
  author={Oquab, Maxime and Darcet, Timoth{\'e}e and Moutakanni, Th{\'e}o and Vo, Huy and Szafraniec, Marc and Khalidov, Vasil and Fernandez, Pierre and Haziza, Daniel and Massa, Francisco and El-Nouby, Alaaeldin and others},
  journal={arXiv preprint arXiv:2304.07193},
  year={2023}
}

@inproceedings{peebles2023scalable,
  title={Scalable diffusion models with transformers},
  author={Peebles, William and Xie, Saining},
  booktitle={Proceedings of the IEEE/CVF international conference on computer vision},
  pages={4195--4205},
  year={2023}
}

@article{albergo2022building,
  title={Building normalizing flows with stochastic interpolants},
  author={Albergo, Michael S and Vanden-Eijnden, Eric},
  journal={arXiv preprint arXiv:2209.15571},
  year={2022}
}

@article{cheang2024gr,
  title={Gr-2: A generative video-language-action model with web-scale knowledge for robot manipulation},
  author={Cheang, Chi-Lam and Chen, Guangzeng and Jing, Ya and Kong, Tao and Li, Hang and Li, Yifeng and Liu, Yuxiao and Wu, Hongtao and Xu, Jiafeng and Yang, Yichu and others},
  journal={arXiv preprint arXiv:2410.06158},
  year={2024}
}

@article{kim2024openvla,
  title={Openvla: An open-source vision-language-action model},
  author={Kim, Moo Jin and Pertsch, Karl and Karamcheti, Siddharth and Xiao, Ted and Balakrishna, Ashwin and Nair, Suraj and Rafailov, Rafael and Foster, Ethan and Lam, Grace and Sanketi, Pannag and others},
  journal={arXiv preprint arXiv:2406.09246},
  year={2024}
}

@article{team2024octo,
  title={Octo: An open-source generalist robot policy},
  author={Team, Octo Model and Ghosh, Dibya and Walke, Homer and Pertsch, Karl and Black, Kevin and Mees, Oier and Dasari, Sudeep and Hejna, Joey and Kreiman, Tobias and Xu, Charles and others},
  journal={arXiv preprint arXiv:2405.12213},
  year={2024}
}

@article{wen2025tinyvla,
  title={Tinyvla: Towards fast, data-efficient vision-language-action models for robotic manipulation},
  author={Wen, Junjie and Zhu, Yichen and Li, Jinming and Zhu, Minjie and Tang, Zhibin and Wu, Kun and Xu, Zhiyuan and Liu, Ning and Cheng, Ran and Shen, Chaomin and others},
  journal={IEEE Robotics and Automation Letters},
  year={2025},
  publisher={IEEE}
}

@article{black2410pi0,
  title={pi0: A vision-language-action flow model for general robot control. CoRR, abs/2410.24164, 2024. doi: 10.48550},
  author={Black, Kevin and Brown, Noah and Driess, Danny and Esmail, Adnan and Equi, Michael and Finn, Chelsea and Fusai, Niccolo and Groom, Lachy and Hausman, Karol and Ichter, Brian and others},
  journal={arXiv preprint ARXIV.2410.24164}
}

@inproceedings{ha2023scaling,
  title={Scaling up and distilling down: Language-guided robot skill acquisition},
  author={Ha, Huy and Florence, Pete and Song, Shuran},
  booktitle={Conference on Robot Learning},
  pages={3766--3777},
  year={2023},
  organization={PMLR}
}

@article{bai2025qwen2,
  title={Qwen2.5-VL Technical Report},
  author={Bai, Shuai and Chen, Keqin and Liu, Xuejing and Wang, Jialin and Ge, Wenbin and Song, Sibo and Dang, Kai and Wang, Peng and Wang, Shijie and Tang, Jun and others},
  journal={arXiv preprint arXiv:2502.13923},
  year={2025}
}

@inproceedings{chen2024internvl,
  title={Internvl: Scaling up vision foundation models and aligning for generic visual-linguistic tasks},
  author={Chen, Zhe and Wu, Jiannan and Wang, Wenhai and Su, Weijie and Chen, Guo and Xing, Sen and Zhong, Muyan and Zhang, Qinglong and Zhu, Xizhou and Lu, Lewei and others},
  booktitle={Proceedings of the IEEE/CVF conference on computer vision and pattern recognition},
  pages={24185--24198},
  year={2024}
}

@article{chi2023diffusion,
  title={Diffusion policy: Visuomotor policy learning via action diffusion},
  author={Chi, Cheng and Xu, Zhenjia and Feng, Siyuan and Cousineau, Eric and Du, Yilun and Burchfiel, Benjamin and Tedrake, Russ and Song, Shuran},
  journal={The International Journal of Robotics Research},
  pages={02783649241273668},
  year={2023},
  publisher={SAGE Publications Sage UK: London, England}
}

@article{lapo,
  title={Learning to act without actions},
  author={Schmidt, Dominik and Jiang, Minqi},
  journal={arXiv preprint arXiv:2312.10812},
  year={2023}
}

@article{lapa,
  title={Latent action pretraining from videos},
  author={Ye, Seonghyeon and Jang, Joel and Jeon, Byeongguk and Joo, Sejune and Yang, Jianwei and Peng, Baolin and Mandlekar, Ajay and Tan, Reuben and Chao, Yu-Wei and Lin, Bill Yuchen and others},
  journal={arXiv preprint arXiv:2410.11758},
  year={2024}
}

@article{yang2025learning,
  title={CoMo: Learning Continuous Latent Motion from Internet Videos for Scalable Robot Learning},
  author={Yang, Jiange and Shi, Yansong and Zhu, Haoyi and Liu, Mingyu and Ma, Kaijing and Wang, Yating and Wu, Gangshan and He, Tong and Wang, Limin},
  journal={arXiv preprint arXiv:2505.17006},
  year={2025}
}

@article{bu2025univla,
  title={Univla: Learning to act anywhere with task-centric latent actions},
  author={Bu, Qingwen and Yang, Yanting and Cai, Jisong and Gao, Shenyuan and Ren, Guanghui and Yao, Maoqing and Luo, Ping and Li, Hongyang},
  journal={arXiv preprint arXiv:2505.06111},
  year={2025}
}

@article{gao2025adaworld,
  title={Adaworld: Learning adaptable world models with latent actions},
  author={Gao, Shenyuan and Zhou, Siyuan and Du, Yilun and Zhang, Jun and Gan, Chuang},
  journal={arXiv preprint arXiv:2503.18938},
  year={2025}
}

@article{moto,
  title={Moto: Latent motion token as the bridging language for robot manipulation},
  author={Chen, Yi and Ge, Yuying and Li, Yizhuo and Ge, Yixiao and Ding, Mingyu and Shan, Ying and Liu, Xihui},
  journal={arXiv preprint arXiv:2412.04445},
  year={2024}
}

@inproceedings{esser2024scaling,
  title={Scaling rectified flow transformers for high-resolution image synthesis},
  author={Esser, Patrick and Kulal, Sumith and Blattmann, Andreas and Entezari, Rahim and M{\"u}ller, Jonas and Saini, Harry and Levi, Yam and Lorenz, Dominik and Sauer, Axel and Boesel, Frederic and others},
  booktitle={Forty-first international conference on machine learning},
  year={2024}
}

@inproceedings{flow,
  title={Flow as the Cross-domain Manipulation Interface},
  author={Xu, Mengda and Xu, Zhenjia and Xu, Yinghao and Chi, Cheng and Wetzstein, Gordon and Veloso, Manuela and Song, Shuran},
  booktitle={CoRL}
}

@article{atm,
  title={Any-point trajectory modeling for policy learning},
  author={Wen, Chuan and Lin, Xingyu and So, John and Chen, Kai and Dou, Qi and Gao, Yang and Abbeel, Pieter},
  journal={arXiv preprint arXiv:2401.00025},
  year={2023}
}

@article{gr2,
  title={Gr-2: A generative video-language-action model with web-scale knowledge for robot manipulation},
  author={Cheang, Chi-Lam and Chen, Guangzeng and Jing, Ya and Kong, Tao and Li, Hang and Li, Yifeng and Liu, Yuxiao and Wu, Hongtao and Xu, Jiafeng and Yang, Yichu and others},
  journal={arXiv preprint arXiv:2410.06158},
  year={2024}
}

@article{gen2act,
  title={Gen2act: Human video generation in novel scenarios enables generalizable robot manipulation},
  author={Bharadhwaj, Homanga and Dwibedi, Debidatta and Gupta, Abhinav and Tulsiani, Shubham and Doersch, Carl and Xiao, Ted and Shah, Dhruv and Xia, Fei and Sadigh, Dorsa and Kirmani, Sean},
  journal={arXiv preprint arXiv:2409.16283},
  year={2024}
}

@article{egovid,
  title={EgoVid-5M: A Large-Scale Video-Action Dataset for Egocentric Video Generation},
  author={Wang, Xiaofeng and Zhao, Kang and Liu, Feng and Wang, Jiayu and Zhao, Guosheng and Bao, Xiaoyi and Zhu, Zheng and Zhang, Yingya and Wang, Xingang},
  journal={arXiv preprint arXiv:2411.08380},
  year={2024}
}

@article{droid,
  title={Droid: A large-scale in-the-wild robot manipulation dataset},
  author={Khazatsky, Alexander and Pertsch, Karl and Nair, Suraj and Balakrishna, Ashwin and Dasari, Sudeep and Karamcheti, Siddharth and Nasiriany, Soroush and Srirama, Mohan Kumar and Chen, Lawrence Yunliang and Ellis, Kirsty and others},
  journal={arXiv preprint arXiv:2403.12945},
  year={2024}
}

@article{libero,
  title={Libero: Benchmarking knowledge transfer for lifelong robot learning},
  author={Liu, Bo and Zhu, Yifeng and Gao, Chongkai and Feng, Yihao and Liu, Qiang and Zhu, Yuke and Stone, Peter},
  journal={Advances in Neural Information Processing Systems},
  volume={36},
  pages={44776--44791},
  year={2023}
}

@inproceedings{genie,
  title={Genie: Generative interactive environments},
  author={Bruce, Jake and Dennis, Michael D and Edwards, Ashley and Parker-Holder, Jack and Shi, Yuge and Hughes, Edward and Lai, Matthew and Mavalankar, Aditi and Steigerwald, Richie and Apps, Chris and others},
  booktitle={ICML},
  year={2024}
}

@article{vla_improve,
  title={Fine-tuning vision-language-action models: Optimizing speed and success},
  author={Kim, Moo Jin and Finn, Chelsea and Liang, Percy},
  journal={arXiv preprint arXiv:2502.19645},
  year={2025}
}

@inproceedings{openx,
  title={Open x-embodiment: Robotic learning datasets and rt-x models: Open x-embodiment collaboration 0},
  author={O’Neill, Abby and Rehman, Abdul and Maddukuri, Abhiram and Gupta, Abhishek and Padalkar, Abhishek and Lee, Abraham and Pooley, Acorn and Gupta, Agrim and Mandlekar, Ajay and Jain, Ajinkya and others},
  booktitle={2024 IEEE International Conference on Robotics and Automation (ICRA)},
  pages={6892--6903},
  year={2024},
  organization={IEEE}
}

@article{survey,
  title={Towards generalist robot learning from internet video: A survey},
  author={McCarthy, Robert and Tan, Daniel CH and Schmidt, Dominik and Acero, Fernando and Herr, Nathan and Du, Yilun and Thuruthel, Thomas G and Li, Zhibin},
  journal={arXiv preprint arXiv:2404.19664},
  year={2024}
}

@article{agibot,
  title={Agibot world colosseo: A large-scale manipulation platform for scalable and intelligent embodied systems},
  author={Bu, Qingwen and Cai, Jisong and Chen, Li and Cui, Xiuqi and Ding, Yan and Feng, Siyuan and Gao, Shenyuan and He, Xindong and Huang, Xu and Jiang, Shu and others},
  journal={arXiv preprint arXiv:2503.06669},
  year={2025}
}

@article{gr00t,
  title={Gr00t n1: An open foundation model for generalist humanoid robots},
  author={Bjorck, Johan and Casta{\~n}eda, Fernando and Cherniadev, Nikita and Da, Xingye and Ding, Runyu and Fan, Linxi and Fang, Yu and Fox, Dieter and Hu, Fengyuan and Huang, Spencer and others},
  journal={arXiv preprint arXiv:2503.14734},
  year={2025}
}

@article{alain2016understanding,
  title={Understanding intermediate layers using linear classifier probes},
  author={Alain, Guillaume and Bengio, Yoshua},
  journal={arXiv preprint arXiv:1610.01644},
  year={2016}
}

@article{beyer2024paligemma,
  title={Paligemma: A versatile 3b vlm for transfer},
  author={Beyer, Lucas and Steiner, Andreas and Pinto, Andr{\'e} Susano and Kolesnikov, Alexander and Wang, Xiao and Salz, Daniel and Neumann, Maxim and Alabdulmohsin, Ibrahim and Tschannen, Michael and Bugliarello, Emanuele and others},
  journal={arXiv preprint arXiv:2407.07726},
  year={2024}
}

@article{wang2025unified,
  title={Unified Vision-Language-Action Model},
  author={Wang, Yuqi and Li, Xinghang and Wang, Wenxuan and Zhang, Junbo and Li, Yingyan and Chen, Yuntao and Wang, Xinlong and Zhang, Zhaoxiang},
  journal={arXiv preprint arXiv:2506.19850},
  year={2025}
}

@article{zhang2025dreamvla,
  title={DreamVLA: a vision-language-action model dreamed with comprehensive world knowledge},
  author={Zhang, Wenyao and Liu, Hongsi and Qi, Zekun and Wang, Yunnan and Yu, Xinqiang and Zhang, Jiazhao and Dong, Runpei and He, Jiawei and Wang, He and Zhang, Zhizheng and others},
  journal={arXiv preprint arXiv:2507.04447},
  year={2025}
}

@article{cen2025worldvla,
  title={WorldVLA: Towards Autoregressive Action World Model},
  author={Cen, Jun and Yu, Chaohui and Yuan, Hangjie and Jiang, Yuming and Huang, Siteng and Guo, Jiayan and Li, Xin and Song, Yibing and Luo, Hao and Wang, Fan and others},
  journal={arXiv preprint arXiv:2506.21539},
  year={2025}
}

@article{li2025unified,
  title={Unified video action model},
  author={Li, Shuang and Gao, Yihuai and Sadigh, Dorsa and Song, Shuran},
  journal={arXiv preprint arXiv:2503.00200},
  year={2025}
}

@inproceedings{radosavovic2023robot,
  title={Robot learning with sensorimotor pre-training},
  author={Radosavovic, Ilija and Shi, Baifeng and Fu, Letian and Goldberg, Ken and Darrell, Trevor and Malik, Jitendra},
  booktitle={Conference on Robot Learning},
  pages={683--693},
  year={2023},
  organization={PMLR}
}

@article{nair2022r3m,
  title={R3m: A universal visual representation for robot manipulation},
  author={Nair, Suraj and Rajeswaran, Aravind and Kumar, Vikash and Finn, Chelsea and Gupta, Abhinav},
  journal={arXiv preprint arXiv:2203.12601},
  year={2022}
}

@article{xiao2022masked,
  title={Masked visual pre-training for motor control},
  author={Xiao, Tete and Radosavovic, Ilija and Darrell, Trevor and Malik, Jitendra},
  journal={arXiv preprint arXiv:2203.06173},
  year={2022}
}

@article{majumdar2023we,
  title={Where are we in the search for an artificial visual cortex for embodied intelligence?},
  author={Majumdar, Arjun and Yadav, Karmesh and Arnaud, Sergio and Ma, Jason and Chen, Claire and Silwal, Sneha and Jain, Aryan and Berges, Vincent-Pierre and Wu, Tingfan and Vakil, Jay and others},
  journal={Advances in Neural Information Processing Systems},
  volume={36},
  pages={655--677},
  year={2023}
}

@article{ye2024latent,
  title={Latent action pretraining from videos},
  author={Ye, Seonghyeon and Jang, Joel and Jeon, Byeongguk and Joo, Sejune and Yang, Jianwei and Peng, Baolin and Mandlekar, Ajay and Tan, Reuben and Chao, Yu-Wei and Lin, Bill Yuchen and others},
  journal={arXiv preprint arXiv:2410.11758},
  year={2024}
}

@article{gao2024flip,
  title={Flip: Flow-centric generative planning as general-purpose manipulation world model},
  author={Gao, Chongkai and Zhang, Haozhuo and Xu, Zhixuan and Cai, Zhehao and Shao, Lin},
  journal={arXiv preprint arXiv:2412.08261},
  year={2024}
}

@article{xu2024flow,
  title={Flow as the cross-domain manipulation interface},
  author={Xu, Mengda and Xu, Zhenjia and Xu, Yinghao and Chi, Cheng and Wetzstein, Gordon and Veloso, Manuela and Song, Shuran},
  journal={arXiv preprint arXiv:2407.15208},
  year={2024}
}

@article{li2025bridgevla,
  title={BridgeVLA: Input-Output Alignment for Efficient 3D Manipulation Learning with Vision-Language Models},
  author={Li, Peiyan and Chen, Yixiang and Wu, Hongtao and Ma, Xiao and Wu, Xiangnan and Huang, Yan and Wang, Liang and Kong, Tao and Tan, Tieniu},
  journal={arXiv preprint arXiv:2506.07961},
  year={2025}
}

@article{qu2025spatialvla0,
  title     = {SpatialVLA: Exploring Spatial Representations for Visual-Language-Action Model},
  author    = {Delin Qu and Haoming Song and Qizhi Chen and Yuanqi Yao and Xinyi Ye and Yani Ding and Zhigang Wang and Jiayuan Gu and Bin Zhao and Dong Wang and Xuelong Li},
  journal   = {ROBOTICS},
  year      = {2025},
  doi       = {10.48550/arXiv.2501.15830},
  bibSource = {Semantic Scholar https://www.semanticscholar.org/paper/10301766e5686bda76722ef2af1362213b934cc0}
}

@inproceedings{zhao2025cot,
  title={Cot-vla: Visual chain-of-thought reasoning for vision-language-action models},
  author={Zhao, Qingqing and Lu, Yao and Kim, Moo Jin and Fu, Zipeng and Zhang, Zhuoyang and Wu, Yecheng and Li, Zhaoshuo and Ma, Qianli and Han, Song and Finn, Chelsea and others},
  booktitle={Proceedings of the Computer Vision and Pattern Recognition Conference},
  pages={1702--1713},
  year={2025}
}

@article{pertsch2025fast,
  title   = {FAST: Efficient Action Tokenization for Vision-Language-Action Models},
  author  = {Karl Pertsch and Kyle Stachowicz and Brian Ichter and Danny Driess and Suraj Nair and Quan Vuong and Oier Mees and Chelsea Finn and Sergey Levine},
  year    = {2025},
  journal = {arXiv preprint arXiv: 2501.09747}
}

@inproceedings{gu2023maniskill2,
  title={ManiSkill2: A Unified Benchmark for Generalizable Manipulation Skills},
  author={Gu, Jiayuan and Xiang, Fanbo and Li, Xuanlin and Ling, Zhan and Liu, Xiqiang and Mu, Tongzhou and Tang, Yihe and Tao, Stone and Wei, Xinyue and Yao, Yunchao and Yuan, Xiaodi and Xie, Pengwei and Huang, Zhiao and Chen, Rui and Su, Hao},
  booktitle={International Conference on Learning Representations},
  year={2023}
}

@article{zhou2024dino,
  title={Dino-wm: World models on pre-trained visual features enable zero-shot planning},
  author={Zhou, Gaoyue and Pan, Hengkai and LeCun, Yann and Pinto, Lerrel},
  journal={arXiv preprint arXiv:2411.04983},
  year={2024}
}

@article{liu2025bridge,
  title={Bridge Thinking and Acting: Unleashing Physical Potential of VLM with Generalizable Action Expert},
  author={Liu, Mingyu and Huang, Zheng and Lin, Xiaoyi and Zhu, Muzhi and Zhao, Canyu and Du, Zongze and Wang, Yating and Zhu, Haoyi and Chen, Hao and Shen, Chunhua},
  journal={arXiv preprint arXiv:2510.03896},
  year={2025}
}

@inproceedings{wang2025vq,
  title={Vq-vla: Improving vision-language-action models via scaling vector-quantized action tokenizers},
  author={Wang, Yating and Zhu, Haoyi and Liu, Mingyu and Yang, Jiange and Fang, Hao-Shu and He, Tong},
  booktitle={Proceedings of the IEEE/CVF International Conference on Computer Vision},
  pages={11089--11099},
  year={2025}
}

@inproceedings{wang2026odyssey,
  title={Odyssey: Open-world quadrupeds exploration and manipulation for long-horizon tasks},
  author={Wang, Kaijun and Lu, Liqin and Liu, Mingyu and Jiang, Jianuo and Li, Zeju and Zhang, Bolin and Zheng, Wancai and Yu, Xinyi and Chen, Hao and Shen, Chunhua},
  booktitle={Proceedings of the AAAI Conference on Artificial Intelligence},
  volume={40},
  number={22},
  pages={18602--18610},
  year={2026}
}
}


\end{document}